\documentclass[12pt]{article}
\usepackage[utf8]{inputenc}
\usepackage{enumitem}
\usepackage{fullpage}
\usepackage{authblk}
\usepackage{amsmath}
\usepackage{graphicx}
\usepackage{hyperref}
\usepackage[maxbibnames=99, doi=false, isbn=false, url=false, eprint=false]{biblatex}
\title{Arbitrarily Large Labelled Random Satisfiability Formulas for Machine Learning Training\footnote{Authors are listed in alphabetical order. Corresponding author: Amrit Daswaney (amritdaswaney@gmail.com)}}
\date{}
\author[1]{\small Dimitris Achlioptas}
\author[2]{\small Amrit Daswaney}
\author[3]{\small Periklis A. Papakonstantinou}
\affil[1]{\footnotesize University of Athens}
\affil[2]{\footnotesize Frogdata}
\affil[3]{\footnotesize Rutgers University}

\usepackage{caption}
\usepackage{subcaption}

\begin{document}

\maketitle

\begin{abstract}
Machine learning methods hold great promise for solving real-life instances of combinatorial problems. Research in this vein has largely focused on the Boolean satisfiability (SAT) problem, due to its theoretical importance and practical applications. A major barrier faced is that while real-life formulas typically  have several thousands of variables, training sets are limited to random formulas with a few hundred variables since labeling larger random formulas is intractable. Crucially, this restriction in the size of training formulas restricts the scale of the models that can be meaningfully trained, creating a highly undesirable situation. We remove this barrier completely by showing how to generate balanced training sets of correctly labeled random formulas of any desired size efficiently, i.e., without solving the underlying decision problem for each formula in the set. Moreover, the difficulty of the classification task for the formulas we generate is tunable via a single scalar parameter, allowing for the exploration and development of a wide range of machine learning models. 
\end{abstract}

\section{Introduction}

Machine learning has been successfully applied to a wide range of domains where the instance distribution is not understood and, thus, no clear mathematical formulation of the problem can be constructed~\cite{prouvost-2021}. Recently, there has been great interest in investigating whether it can  be applied to combinatorial problems. 

Central in this effort is  Boolean Satisfiability (SAT), the canonical \textsf{NP}-complete problem~\cite{cook-1971}. Besides its immense theoretical importance, SAT has several practical applications, ranging from hardware and software verification to planning and scheduling~\cite{selman-2008}. So far, researchers have built machine learning methods to predict satisfiability~\cite{dill-2019a, leyton-brown-2020} and/or to find a satisfying assignment~\cite{dill-2019a, weimer-2019}.

A major barrier in the effort to use machine learning for satisfiability is the absence of datasets comprising large labeled formulas from distributions for which the decision problem appears hard. Specifically, in order to form a (balanced) training set, all existing works generate random \mbox{3-CNF} formulas ``at the threshold,'' so that the probability of satisfiability is (approximately) 1/2, and then use a complete (DPLL) solver to decide whether each generated formula is satisfiable or not. Since resolution proofs of unsatisfiability (and thus DPLL solver executions) for random formulas are exponentially large in the number of variables~\cite{chvatal-1988}, this means that only relatively small formulas can be labeled, i.e., orders of magnitude smaller than practically relevant formulas. To make things worse, the satisfiable formulas generated in this manner are very easy for modern SAT solvers, much unlike their unsatisfiable counterparts~\cite{selman-2008}. 

The use of random formulas as above for training  raises serious concerns about the generalization of the trained models to practically relevant formulas. This is due not only to the small size of the formulas in the training set but also to the fragility of the generative model, as we need to be ``right at the threshold'' in order to get a balanced data set. Concomitantly, the state-of-the-art models for classifying such formulas~\cite{dill-2019a, leyton-brown-2020}, having been developed around formulas of small size, appear intractable to train on formulas of practically interesting size, as their training time increases too rapidly with formula size.

In this work, we exploit the Probabilistic Method~\cite{alon-1992} to introduce a generator that overcomes all of the aforementioned concerns.  Specifically, our generator can efficiently generate arbitrarily large, labeled random $k$-CNF formulas, enabling researchers to train models on any desired problem size. Our generator has the additional property that the difficulty of finding a satisfying truth assignment for its satisfiable formulas can be modulated from relatively easy to extremely hard. Finally, and perhaps most importantly, our generator makes it possible to consider a new, more robust, learning task, wherein instead of distinguishing between satisfiable and unsatisfiable formulas, we need to distinguish between formulas that are satisfiable and formulas that are far from satisfiable, i.e., formulas where every truth assignment violates a significant number of clauses.

\subsection{Overview}

Recall that to generate a random 3-CNF formula with $n$ variables and $m$ clauses one simply selects uniformly and independently $m$ clauses out of all $8 \binom{n}{3}$ possible 3-clauses on the $n$ variables. The interesting asymptotic regime is when $m = r n$, i.e. when $m$ is linear in $n$. The satisfiability threshold is conjectured to occur around density $r \approx 4.267...$~\cite{zecchina-2005a}. As mentioned, the method used so far to generate training examples is to generate a uniformly random 3-CNF formula of density $r=4.267$ and a few hundred variables and then use a complete DPLL solver to determine whether it is satisfiable or not. Naturally, the second step crucially curtails the size of formulas that can be labeled. 

Since certifying that a given random 3-CNF formula is satisfiable is widely considered an intractable task, our key insight is to sidestep it completely, by replacing absolute certainty (of unsatisfiability) with virtual certainty, i.e., overwhelming probability. The reason we can get do this is that for densities not far above the satisfiability threshold, as we will see, the probability that a random formula is satisfiable is provably \emph{exponentially} small in its number of variables. As a result, increasing formula size plays \emph{in our favor} and our generator never invokes a satisfiability solver, proceeding instead as follows.

First, a density $r^*$ above the satisfiability threshold is fixed, e.g., $r^*=5$. The exact value of $r^*$ is not very important, making the process robust. Then, if $n$ is the desired number of formula variables and $C$ is the set of all $8 \binom{n}{3}$ possible clauses on $n$ variables, we proceed to select $m=r^*n$ clauses with 3 literals independently, as follows:
\begin{enumerate}
\item[(a)]
To generate a negative (unsatisfiable) example, clauses are selected \emph{uniformly} from $C$.
\item[(b)]
To generate a positive (satisfiable) example, first, a truth assignment $A \in \{0,1\}^n$ is selected uniformly. If $C_A$ is the set of  $7 \binom{n}{3}$ clauses satisfied by $A$, then each clause in $C_A$ is selected with probability proportional to $q^t$, where $t \ge 1$ is its number of satisfied literals under $A$ and $q < 1$ serves as the hardness-controlling parameter.
\end{enumerate}

A priori, it is not clear why either of the methods (a,b) above serves its stated purpose. In a nutshell, the idea in each case is as follows. 

(a) We will see that there exists a density $r_c \approx 4.667$, such that if a random 3-CNF formula has density $r = r_c + \delta$, then the probability that it is satisfiable is $\exp(-\Theta(\delta n)))$. As a result, if, for example, we form a dataset with, say, $100,000$ formulas, each having $10,000$ variables and $50,000$ clauses, the probability that \emph{even a single one} of them will be satisfiable is approximately $10^{-148}$. While we give an explicit mathematical formula for this probability as a function of all relevant quantities, it should suffice to say that for all $r^* \ge 5$ and $n \ge 10,000$, one does not really need to bother with the computation, as the probability of generating any satisfiable formula is simply astronomically small. 

(b) To generate \emph{satisfiable} formulas of density above the satisfiability threshold, e.g., at $r^* =5$, we need to ``plant'' a random satisfying assignment, $A$, which is precisely what we do by never selecting clauses violated by $A$. Such naive planting, though, is not enough: if we were to sample the clauses satisfied by $A$ uniformly, i.e., with equal probability, the planted assignment would be easy to find~\cite{moore-2005a} and the formulas would be easily distinguished from the uniformly random formulas of the same density that serve as our unsatisfiable formulas.

To overcome the problems associated with naive planting, we leverage the work of~\cite{strain-2005a}, inspired by the work of~\cite{peres-2003}. The intuitive idea is that by taking $q < 1$, i.e., by selecting clauses with many satisfied literals with lower probability, the bias of each variable towards its value in $A$ is reduced, to the point that for small enough $q$ the bias is, in fact, in the direction of $\bar{A}$, deceiving local search solvers. Combined with the fact that, due to the high density, the only satisfying assignments are $A$ and, perhaps, a few more in its vicinity, hardness emerges. Indeed, as was shown in~\cite{strain-2005a} (and as we reproduce here), for each density $r^*$, there is a relatively wide range of values of $q$, such that the resulting formulas are hard. \medskip

Having introduced a new generator of labeled formulas, we must of course demonstrate that the classification task it induces is learnable, for otherwise, it would not be of much value. To do this we started off by generating a balanced dataset of satisfiable and unsatisfiable formulas with 10,000 variables each. Confirming our concerns about the generalization of existing state-of-the-art models~\cite{dill-2019a, leyton-brown-2020}, we found that they perform \emph{no better than random guessing} on our formulas, both when used ``off the shelf,'' i.e., when trained on random formulas at the threshold, and when trained using examples from our generator.

Our second contribution is to present a novel classifier that performs significantly better than random guessing (99\%) on the same datasets for most values of $q$ (difficulty levels). In contrast to existing classifiers that rely on syntactic features of the formula to be classified, our classifier makes its decision on the basis of statistics concerning a short prefix of a solver's computation on the formula. We believe that using such statistics to drive machine learning models for satisfiability is an exciting new area, as demonstrated by the fact that our classifier works well not only on the formulas of our generator but also on formulas ``at the threshold,'' indeed improving upon the state of the art.

\section{Prior Work}

\textbf{Problem generators.} The machine learning community has used uniform random $k$-SAT formulas as training datasets~\cite{shoham-2009, leyton-brown-2007, leyton-brown-2012, dill-2019a, leyton-brown-2020}. However, in the SAT community, other formula generators are also used (see ~\cite{levy-2015} and references therein). The common idea is to identify some practically relevant class of formulas and use real-life instances from that class to train a model that outputs new formulas that resemble the real-life formulas. The original class can often be characterized by a small set of properties of interest. For example, researchers have found that the graph representation of several industrial formulas has ``community structure.'' Informally, a graph has a community structure, if it can be decomposed into subgraphs that have more internal edges than outgoing edges. Each subgraph is called a community. The aim would then be to create new formulas that also have community structure~\cite{levy-2015}.

Unfortunately, these formula generators face the same problem outlined for uniform random formulas, namely they need labeled examples in order to train the generative model, something which, in turn, requires invoking a SAT solver. This limits the size of the formulas to a few thousand variables. Additionally, existing experimental evidence~\cite{ansotegui-2009, levy-2015, levy-2017} suggests that these instances are easier to solve than the class of formulas they aim to resemble. In contrast, we are able to tune our generator to output formulas that range from easy to seemingly extremely hard by changing the value of $q$. As we will see, our experiments suggest that balanced deceptive formulas, i.e., where $q$ is chosen so that the local bias of each variable points neither towards its value in $A$ nor towards the opposite value, is as hard as unsatisfiable formulas.

\textbf{Satisfiability prediction.} 
Initial work on satisfiability prediction was done as part of the development of empirical hardness models~\cite{shoham-2009}. They trained a random forest classifier on the syntactic features of the formula (e.g. number of positive and negative literals) and statistics about variables in the relaxation of the problem (e.g. coefficient of variation of the slack variable in the LP relaxation of the problem) to predict the satisfiability for random 3-SAT formulas with $100, 150, \cdots, 600$ variables~\cite{leyton-brown-2007, leyton-brown-2012}

More recently, \cite{dill-2019a} initiated the study of satisfiability prediction using neural networks. They represent a formula using its literal-clause graph and use a graphical neural network (called `NeuroSAT') with a message-passing architecture to distinguish a pair of satisfiable and unsatisfiable formulas that differ in 1 clause~\cite{dill-2019a}. They consider formulas with 40 to 200 variables. There has been significant follow-up work to NeuroSAT. The most relevant of which is~\cite{leyton-brown-2020}. Using tensors to represent the formula symbolically, they apply NeuroSAT's message-passing architecture and the exchangeable architecture of~\cite{ravanbakhsh-2018a} to predict the satisfiability of random 3-SAT formulas with $100, 200, \cdots, 600$ variables. We closely compare our work to these two prior papers.

\section{Problem Generator} \label{problem-generator}

The core challenge in generating a balanced dataset of large 
formulas is obtaining formulas that can be safely labeled unsatisfiable. This is because labeling formulas as satisfiable is relatively easier, as any satisfying assignment serves as a certificate. The key observation underlying our method is that formulas with clause density above the threshold are unsatisfiable with overwhelming probability. Specifically, let 
\begin{equation}
    f(r,n) = \left(\frac{7}{8}\right)^r \cdot \left( 2 - \left(1 - \frac{3}{7n} \right)^{rn}   \right)  \approx \left(\frac{7}{8}\right)^r \cdot \left( 2 - \left(1 - e^{-3r/7} \right)\right) \enspace, \label{eq1}
\end{equation}
where $r$ is the clause density and $n$ is the number of variables. 

In~\cite{kirousis-1998} it was shown that the probability that a random 3-CNF formula with $n$ variables and $m=rn$ clauses is satisfiable is bounded by $f(r,n)^n$. We use this concrete bound to argue that the formulas we generate are unsatisfiable with overwhelming probability. To see this, observe that $f$ is decreasing in $r$ and crosses 1 around $r \approx 4.667$. In particular, we see that $f(5,n) \approx 0.966$, implying that a random 3-CNF formula with $10,000$ variables and $50,000$ clauses is satisfiable with probability less than $10^{-152}$. Thus, the probability that a data set with a million such formulas will contain even a single satisfiable one is smaller than the probability of a massive meteorite striking Earth in the next minute. 

To obtain satisfiable formulas of the same density as our unsatisfiable formulas (for otherwise the two classes of formulas could be trivially distinguished by a classifier based merely on density) we plant a satisfying assignment in a ``deceptive'' way, as follows:

\begin{enumerate}
    \item Fix a constant $q < 1$.
    \item Generate a random truth assignment $A \in \{0, 1\}^n$.
    \item Repeat until you have $m$ clauses: 
    	\begin{itemize}
	\item
	Pick a clause $c$ uniformly among all $8 {n \choose 3}$ possible clauses on $n$ variables. 
	\item
	If $c$ is satisfied by $A$, then add $c$ to the formula with probability $q^t$, where $t$ is the number of satisfied literals in $c$ under $A$.
	\end{itemize}
\end{enumerate}

As mentioned, the smaller the value of $q$, the smaller the probability of including in the formula clauses with many satisfied literals under $A$. As a result, as $q$ decreases, so does the probability that the ``majority vote'' for each variable coincides with its value under $A$. In particular, the crossover point is $q^* = 0.618...$, for which the average number of satisfied literals per clause under $A$ equals 3/2, i.e., the same as for a random assignment (even though every clause has at least one satisfied literal under $A$). Reducing $q$ below $q^*$, causes the variables to point away from their assignment in $A$, thus creating a ``deceptive'' formula. 

We note that our instance generator has a number of desirable properties: (a) the values of $n$, and $m/n$ can  be freely chosen which allows the asymptotic behavior of solvers to be studied, (b) a large number of instances can be easily and efficiently generated, and (c) the difficulty of test cases can be varied using a single parameter.

\section{Computation-based satisfiability predictor}

In this section, we describe how we use features from a short prefix of an SLS solver's computation to represent a problem instance. Prior work in this area has captured a solver's computation ``externally" by recording variables the authors hypothesized were important. For example, the maximum number of variables set to true~\cite{chickering-2001} and the number of steps to the best local minimum~\cite{shoham-2004}. 
Instead of trying to invent external features that capture problem-solving progress, our key insight is to directly record the solver's internal view of the problem. The intuition is that given the success of these solvers, these features are a more concise and useful representation of the search space. We benefit from the clever choices and pruning a solver must perform when deciding how to explore the search space.

There are 2 challenges we faced when implementing such an approach. 
First, selecting the solver to use. We initially decided to pick an algorithm from the SLS family of algorithms due to their success in solving random CNF formulas. We chose WalkSAT~\cite{cohen-1996} due to its simplicity and because it forms the basis of several more modern SLS solvers. WalkSAT starts by randomly selecting a starting assignment and taking steps until all clauses are satisfied, or the maximum number of steps has been reached. In every step, it randomly selects an unsatisfied clause $C$ and checks to see if any variable in $C$ can be flipped without causing any other clause to become unsatisfied. If such a variable exists, it is selected and flipped. Otherwise, the algorithm flips either the variable in $C$ whose flipping causes the fewest number of clauses to become violated or a random variable in $C$. The probability of each of these two possibilities is controlled by the only parameter of the algorithm, denoted by $wp$, modulating how greedy each step is. 

Second, deciding how to represent the solver's computation to a classifier. We choose not to give the entire computation to the classifier as this would necessarily scale with the size of the problem instance. This meant compressing the solver's computation in some fashion. We chose to do this in two ways: (a) using summary statistics to get an overall picture of the solver's computation, and (b) treating the solver's computation as a time series and generating time-related features to capture the trajectory of the solver's path. 

At every step of the solver's computation, we record the value of: (1) the variable flipped and its break value, (2) the number of unsatisfied clauses, (3) the clause chosen, (4) the number of steps since the clause chosen was flipped, and (5) Hamming distance between the current assignment from the starting assignment. For each variable and clause, we additionally calculate their count and period. 

Given a list of values for each aforementioned feature, we generate 2 types of features:
\begin{enumerate}
    \item[(a)]
    \textbf{Summary statistics.} We compute the maximum, minimum, average, median, variance, coefficient of variation, first value, last value, quantile and decile values, and inter-quartile (decile) range. We included quantile and decile values to capture potentially different behavior within the same list of values. 
    \item[(b)] 
    \textbf{Time-series related features.} The authors in~\cite{jones-2019} systematically study time series-related features and their utility on various time series-based tasks. They propose a set of 22 features that: (a) exhibit strong classification performance across a given collection of time-series problems, and (b) are minimally redundant. We use these features to capture the solver's trajectory. We refer the reader to page 14 of~\cite{jones-2019} for a full description of the features. In addition to these features, we also calculate the first and second derivatives of each time series.
\end{enumerate}

\section{Experiments}

The goal of this section is to assess the learnability of the formulas generated by our approach. We first attempted to retrain state-of-the-art satisfiability prediction models using publicly available code and training procedures \cite{leyton-brown-2020, dill-2019a}, but found that they performed no better than random guessing on \emph{all} learning tasks considered in this study. To address this, we developed a novel classifier that leverages a short prefix of WalkSAT's computation on the formula for learning. Our classifier was tested on the learning tasks described in \cite{leyton-brown-2020} and \cite{dill-2019a}, and achieved a significantly higher accuracy (99\%) than random guessing for most difficulty levels, demonstrating the learnability of the formulas generated by our approach. Full details for our experimental setup and code, and our replication efforts are provided in appendices \ref{expt-setup} and \ref{ref-prior-work} respectively.

\subsection{Predicting satisfiability on individual formulas.}

\textbf{Experiment.} We test our classifier on the learning task proposed in~\cite{leyton-brown-2020}. The task is to train a classifier on a balanced dataset of random SAT and UNSAT formulas and then test the classifier on unseen random formulas of the same size. We generate 2000 unsatisfiable formulas uniformly at random and 2000 satisfiable deceptive formulas using our generator. Each formula contains $10,000$ variables. We perform an 80-20 stratified train-test split on this dataset. We train our classifiers on the training set and use the test set to measure model accuracy. To reduce variance and ensure robustness of our results, we repeat this entire process 100 times (with different random train-test splits) and report the average test accuracy. We repeat this experiment for different values of the clause density $r \in \{0.5, 0.7, 0.9\}$ and deceptive parameter $q \in \{0.3, 0.4, 0.5, 0.618\}$ to study their impact on classification accuracy.

\textbf{Result.} We report the classification accuracy in Table \ref{table:1}. Our classifier is able to distinguish deceptive formulas $(q < 0.618)$ from uniformly random unsatisfiable formulas with almost perfect classification accuracy (approximately $99\%$). At first glance, this result may seem to be at odds with the results of \cite{moore-2005a} where they report WalkSAT requiring a larger number of steps on average to find a satisfying assigning as the value of $q$ decreases. The reason for this apparent tension is due to the difference in the focus of the 2 papers. In \cite{moore-2005a}, they focus on creating SAT formulas that are hard for WalkSAT to solve. In contrast, our goal is to create SAT formulas that are difficult to differentiate from uniformly random UNSAT formulas. As $q$ is decreased, the local bias of each variable points away from its value in the planted satisfying assignment thereby making it harder for WalkSAT to find the planted assignment. However, this bias apparently leads to a larger number of unsatisfied clauses on average than the uniform random unsatisfiable formula. Our classifier is able to identify and utilize this difference to distinguish SAT from UNSAT formulas. However, when the deceptive formula is balanced (when $q= 0.618$ for 3-SAT), the local bias of each variable points neither towards its value in the planted satisfying assignment nor towards the opposite value. In this case, our classifier is unable to distinguish the formula from a uniformly random unsatisfiable formula. We leave open the question of whether large balanced deceptive formulas can be \emph{efficiently} distinguished from uniform random unsatisfiable formulas.

\begin{table}[h]
\small
\centering
\begin{tabular}{| c | c | c | c |} 
 \hline
 q / r & $r=5$ & $r = 7$ & $r = 9$  \\ [0.5ex] 
 \hline
 0.3 & 98.99\% (0.07\%) & 99.00\% (0.02\%) & 98.91\% (0.02\%) \\
 \hline
 0.4 & 99.02\% (0.06\%) & 99.03\% (0.05\%) & 99.00\% (0.03\%) \\
 \hline
 0.5 & 98.96\% (0.11\%) & 99.01\% (0.01\%) & 98.98\% (0.18\%) \\
 \hline
 0.618 & 49.90\% (1.89\%) & 49.07\% (2.09\%) & 50.04\% (2.23\%) \\
 \hline
\end{tabular}
\caption{Average classification accuracy $\%$ of our computation-based classifier on distinguishing a uniform random 3-SAT formula and a satisfiable formula with a deceptively planted solution for different values of clause density ($r$) and deceptive parameter ($q$). The mean and standard deviation are reported over 100 different random seeds. All formulas have 10,000 variables.}
\label{table:1}
\end{table}

\subsection{Predicting satisfiability on pairs of formulas.}

\textbf{Experiment.} We extend the learning task proposed by~\cite{dill-2019a} to study the classification of pairs of satisfiable and unsatisfiable formulas that differ in multiple clauses. The original learning task proposed by~\cite{dill-2019a} involved training a classifier to distinguish a pair of satisfiable and unsatisfiable uniformly random formulas that differed in a single clause. We extend this to include pairs of formulas that differ in $u$ clauses. This extension allows us to study how far apart, in terms of clauses, a pair of formulas need to be before we can reliably distinguish them. Additionally, it allows finer-grained assessment and comparison across classifiers. To create each pair of formulas, we use our generator to create a satisfiable formula with $10,000$ variables. Then we create a counterpart with the same clauses except we flip literals in exactly $u$ clauses (chosen uniformly at random) to ensure that the new formula does not satisfy the deceptively planted satisfying assignment. We use this procedure to create 1000 pairs of formulas. We perform an 80-20 stratified train-test split on this dataset. We train our classifiers on the training set and use the test set to measure model accuracy. To reduce variance and ensure robustness of our results, we repeat this entire process 100 times (with different random train-test splits). We define a classifier as successful if its average test accuracy over these 100 runs is higher than a pre-specified threshold accuracy, indicating that it is able to reliably distinguish between formula pairs that differ in $u$ clauses. We set the threshold accuracy to be 70\% to ensure that our results are not affected by random variations in the data. Finally, we use binary search to find the minimum value of $u$ for which our classifier was successful. Full details on how formulas were generated and experimental controls are implemented are provided in appendices \ref{gen-pair} and \ref{expt-controls} respectively.

\begin{figure}
     \centering
     \begin{subfigure}[b]{0.45\textwidth}
         \centering
         \includegraphics[width=\textwidth]{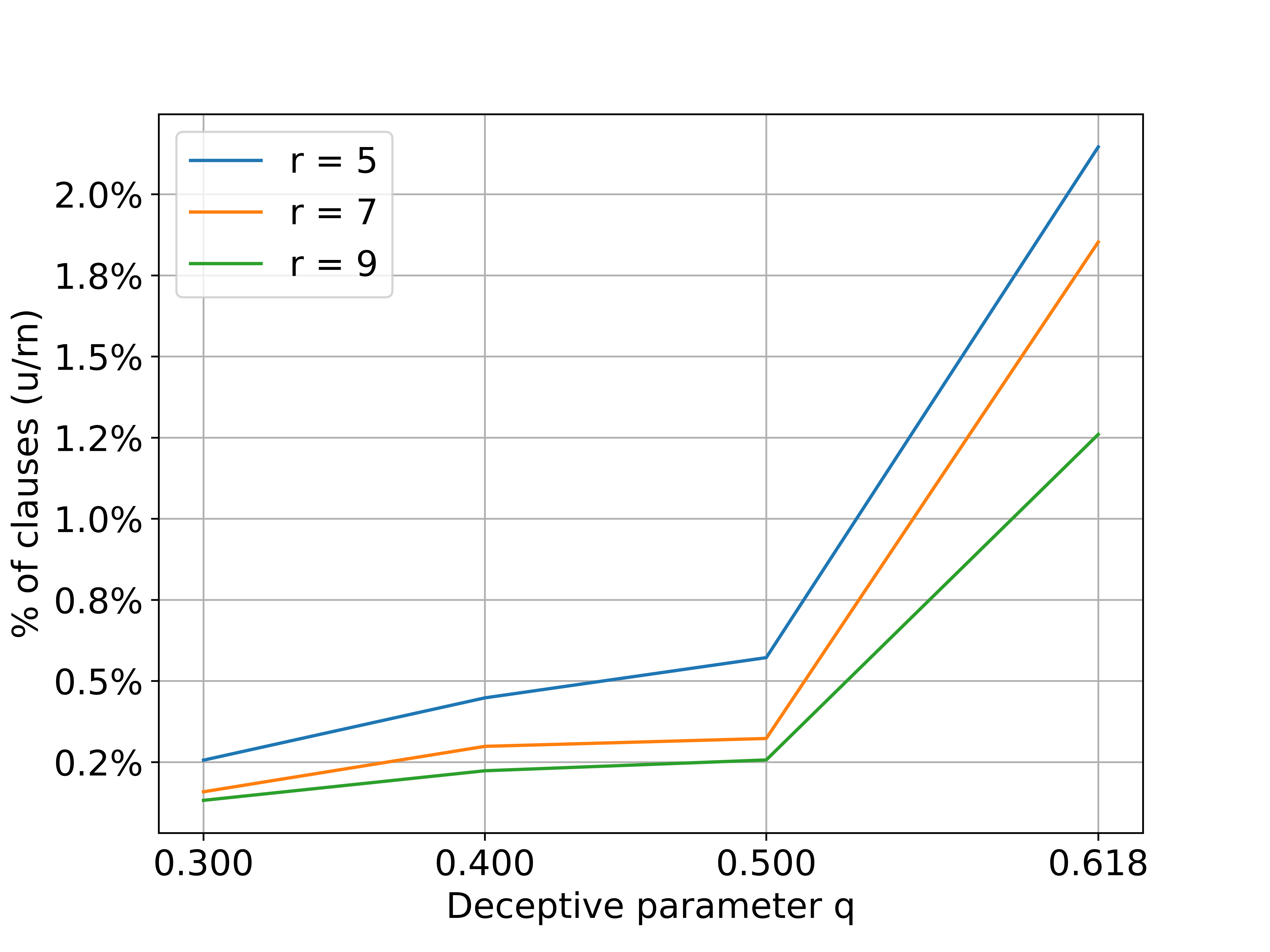}
         \caption{Relative number of clauses}
         \label{fig2a}
     \end{subfigure}
     \begin{subfigure}[b]{0.45\textwidth}
         \centering
         \includegraphics[width=\textwidth]{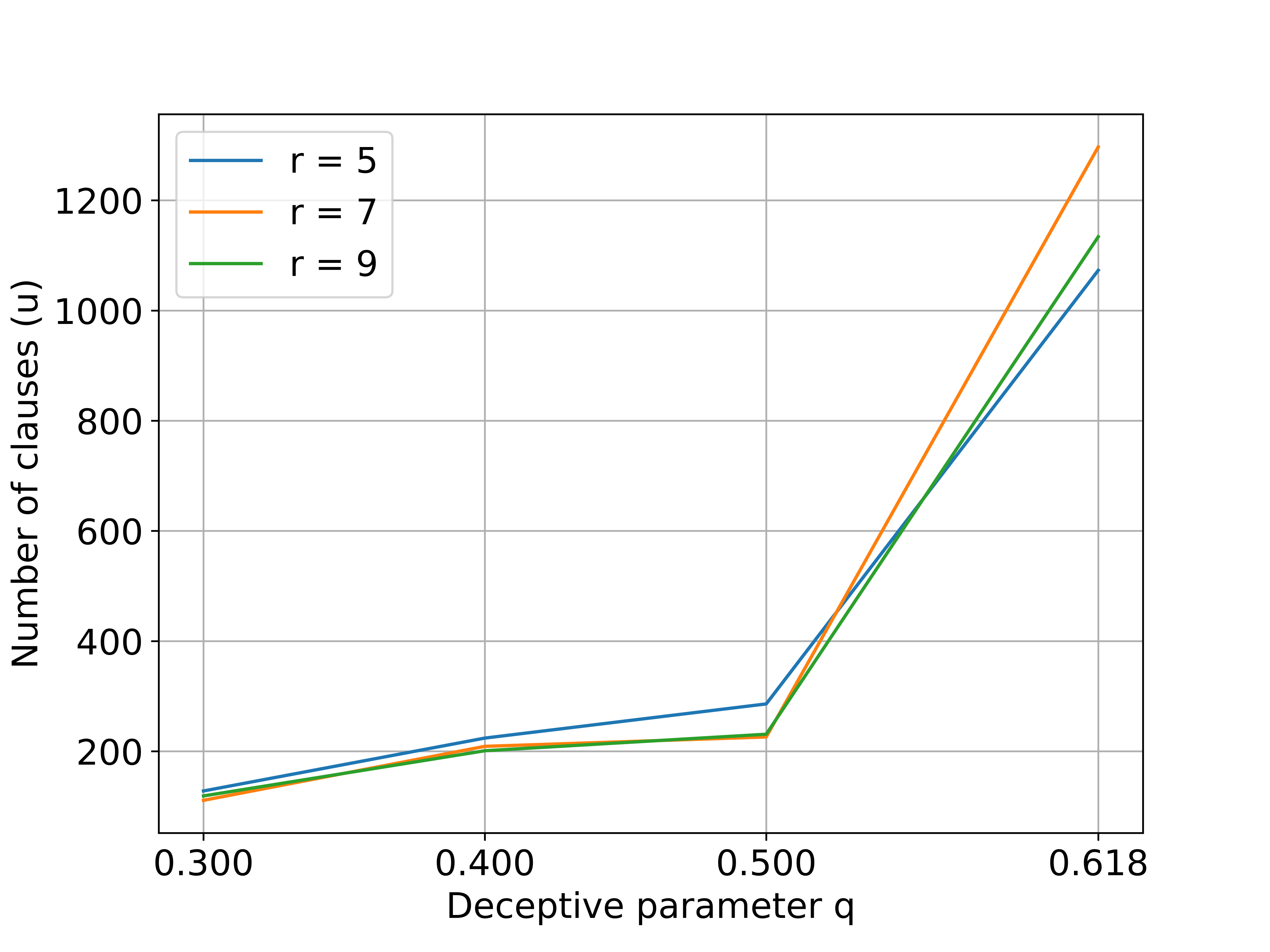}
         \caption{Absolute number of clauses}
         \label{fig2b}
     \end{subfigure}
    \caption{The relative and absolute minimum number of clauses unsatisfied by the deceptively planted satisfying assignment before our classifier can distinguish it from the original satisfiable formula with an average test accuracy \% of at least $70\%$.} 
\end{figure}

\textbf{Result.} Figures \ref{fig2a} and \ref{fig2b} illustrate how the relative and absolute distance (in terms of clauses) required for our classifier to successfully distinguish between a pair of formulas changes as the clause density increases. Our classifier's ability to reliably distinguish \emph{large} deceptive formulas that differ in only 0.1\%-2\% of their clauses is a promising result, as it demonstrates the effectiveness of our approach in identifying subtle differences between formula pairs. An important observation is that both formulas in any pair are initialized with the same random seed, which means that their walks through the search space are initially identical until they select a differing clause. This observation provides important context for understanding the significance of our classifier's performance. Despite the significant increase in the number of clauses in the formulas (ranging from $50-90,000$ clauses), Figure \ref{fig2b} shows that the absolute distance (in terms of clauses) required for our classifier to distinguish between pairs of formulas only modestly increases. This suggests that our classifier is effective at identifying the key differences between formulas and is likely to be able to scale to larger formulas. Additionally, the decrease in the fraction of clauses required as the deceptive parameter increases (as shown in Figure \ref{fig2a}) is accounted for by the modest increase in the value of $u$ required to distinguish between formula pairs. Finally, we observe a steep increase in the fraction of clauses required for the balanced deceptive formulas. This underscores our finding from Experiment 1 that balanced deceptive formulas are hard to distinguish.

\section{Discussion}

One of the ultimate goals of using machine learning on SAT problems is to surpass the existing abilities of modern SAT solvers. Despite the good performance of existing model architectures on small random $k$-CNF formulas, it appears that it is either not possible or intractable to retrain them to do better than random guessing on larger formulas. Given the intractability of correctly labeling UNSAT formulas, it is likely that another approach will be required. Using a sophisticated probabilistic argument, we sidestep this hurdle entirely. Observe, that our bound needed to be exponential in the number of variables, \emph{not} asymptotic, i.e., could be used for finite $n$, and explicit. If any one of the three didn't hold, our generator would not work. Finally, not only do we allow researchers to create datasets containing arbitrarily large, correctly labeled formulas but we allow them to tune the difficulty of these datasets. Observe that training on formulas `at the threshold' is tantamount to trying to learn an extremely fine distinction because in each unsatisfiable formula, removing a handful of clauses makes it satisfiable. In contrast, the added ability to tune the difficulty will allow for more robust and fine-grained comparisons between different models.

\printbibliography

\appendix

\section{Experimental Setup}\label{expt-setup}

\textbf{Hardware and software.} All experiments were conducted on a compute cluster with 688 nodes, each equipped with two 2.60 GHz Intel Xeon 32-bit processors with 2 GB of RAM per processor. All experiments and analyses were written in Python 3.7. Machine learning models from Python's scikit-learn library~\cite{duchesnay-2011} were used. Code to replicate all of our experiments can be accessed here: https://github.com/d-amrit/hide-and-seek.

\medskip

\noindent \textbf{Deceptive formula dataset.} We generated 2000 unsatisfiable random 3-SAT formulas with $10,000$ variables at clause density $r \in \{5, 7, 9\}$. Using our problem generator described in Section \ref{problem-generator}, we create 2000 satisfiable formulas with the same number of variables, clause density, and $q \in \{0.3, 0.4, 0.5, 0.618\}$. We use a range of values for $q$ and $r$ to study their impact on classification accuracy. 

Using \eqref{eq1}, the probability that formulas we label `UNSAT' are actually satisfiable is approximately $0.966^{10000} \approx 1.52 \times 10^{-153}$, $0.766^{10000} \approx 0$, and $0.595^{10000} \approx 0$ for clause densities 5, 7, and 9 respectively. 

\medskip

\noindent \textbf{Choice of SLS solver.} We evaluated our experimental setup on four SLS solvers: WalkSAT~\cite{cohen-1996}, Sch\"{o}ning's algorithm~\cite{schoning-1999}, ProbSAT~\cite{schoning-2012}, and PolyLS~\cite{papakonstantinou-2016}. We used the parameters specified in the respective papers for 3-SAT. We obtained similar results for all models except for ProbSAT with a polynomial function that used both make and break values, which performed no better than random guessing. We present results for WalkSAT as it provided comparable results to ProbSAT and PolyLS and served as the foundation on which the two models were built.

\medskip

\noindent \textbf{Solver parameters.} For each formula with $n$ variables in the dataset, we ran 16 independent instantiations of WalkSAT on the formula with noise parameter (wp) = $0.48$ and max steps set to $2n$. We studied the effect of the number of independent trials, noise parameter value, and max steps parameters on classification accuracy. Increasing the number of independent trials increases the accuracy up to a certain value. We hypothesized that this is because the noise in the feature values due to the randomness of the algorithm decreases as we increase the number of trials. We chose the smallest value of max steps which ensures that WalkSAT can't directly solve the formula and gave us consistent results. The performance of WalkSAT, in terms of runtime distribution, is sensitive to the value of the noise parameter~\cite{selman-2010}. Interestingly, we found that this was \textit{not} the case when performance is measured in terms of classification accuracy. We tested a range of values for the noise parameter and observed similar accuracy across a range of values and the same trend reported below - almost perfect classification accuracy $(\approx 99\%)$ on $q \in {0.3,0.4,0.5}$ and no better than random guessing on $q = 0.618$. We report the findings of the best noise parameter, in terms of solving a formula, $wp = 0.48$ below.

\medskip

\noindent \textbf{Choice of classifier.} For both tasks, we trained the following scikit-learn~\cite{duchesnay-2011} classifiers: Decision Tree, Random Forest, AdaBoost, Gradient Boosting, Histogram Gradient Boosting, Multi-layer Perceptron (MLP). We used the Adam optimizer to train the MLP and tuned the learning rate. Additionally, we trained the XGBoost~\cite{chen-2016}, Cat Boost~\cite{dorogush-2018}, TabNet~\cite{arik-2021a}, LightGBM~\cite{ke-2017}, auto-sklearn~\cite{feurer-2015}, and FLAML~\cite{wang-2021} classifiers. All of the classifiers had similar accuracy on all settings and the same trend reported below - almost perfect classification accuracy $(\approx 99\%)$ on $q \in \{0.3,0.4,0.5\}$ and no better than random guessing on $q = 0.618$. We report the results of a decision tree classifier due to its interpretability.

\medskip

\noindent \textbf{Training procedure.} For all classification tasks, we followed the procedure given below:

\begin{enumerate}
    \itemsep-0.3em
    \item Initialize the random generator with an input seed for replicability.
    \item Perform an 80-20 stratified train-test split on the dataset.
    \item Train a decision tree classifier with max-depth $=3$.
    \item Report the accuracy \% on the test set.
\end{enumerate}

\noindent Repeat the above process for 100 different random seeds and report the average test accuracy \%. 

\medskip

\noindent \textbf{Hyperparameter tuning.} We use scikit-learn's~\cite{duchesnay-2011} grid search optimization to tune the decision tree. For max depth, we consider the values $1, 2, \cdots, 10$ and leaving it unconstrained. For max features, we consider the fractional values $0.1, 0.2, \cdots, 1$, ``sqrt'', ``log2'', and leaving it unconstrained. For minimum samples at the leaf node, we consider the fractional values $0.1, 0.2, \cdots, 1$.

\medskip

\noindent \textbf{Feature Selection.} We use scikit-learn's~\cite{duchesnay-2011} forward recursive feature elimination with stratified 5-fold cross-validation to select the top 3 features.

\section{Comparison to prior work} \label{ref-prior-work}

Using the publicly available code and training procedure specified in their respective papers, we re-trained the models in \cite{leyton-brown-2020} and \cite{dill-2019a} on the same formulas used in our experiment.\footnote{Repositories used for \cite{leyton-brown-2020} and \cite{dill-2019a} are https://github.com/ChrisCameron1/End2EndSAT and https://github.com/dselsam/neurosat respectively}

In addition, for the model used in \cite{leyton-brown-2020}, we corresponded via email with the authors to ensure that we were using the correct training procedure. We trained their model for 72 hours (they use 40 hours in their paper). We found that on \emph{all} datasets, their average test accuracy is no better than 50\%. In addition to testing their model on our formulas, we tested our model on their datasets for formulas with 500, 550, and 600 variables.\footnote{Smaller formulas with $100, 150, 200, \cdots, 450$ are too ``easy'' and WalkSAT was often able to solve them within a few steps.} Using the same training and testing procedure as them, we were able to get accuracy between $86\% - 89\%$, marginally better than the $84\%$ their models are able to achieve.

The model used in \cite{dill-2019a} was trained for $10^3$ iterations for formulas with 200 variables. We re-train their model for $10^6$ iterations for formulas with 10,000 variables. We found that on \emph{all} datasets, their average test accuracy is no better than 50\%.

\section{Generating pairs of formulas for Experiment 2} \label{gen-pair}

In this section, we provide a detailed explanation of how the pairs of formulas for Experiment 2 were generated. Given a formula $F$ with satisfying assignment $A$, we construct a new formula $F_u$ by uniformly randomly selecting $u$ of $F$'s clauses and flipping their literals to ensure that $A$ does not satisfy any of those clauses. We output the pair $(F, F_u)$. The learning task is to identify the formula that $A$ satisfies. 

We create 1000 pairs of formulas with $10,000$ variables using the above process for each pair of values of $q \in \{0.3, 0.4, 0.5, 0.618\}$ and $r \in \{5, 7, 9\}$. We randomly selected 500 pairs of formulas with $F$ in the first slot and $F_u$ in the second slot and the label $y = 0$. We put $F_u$ in the first slot and $F$ in the second slot with the label $y=1$ for the other 500 pairs of formulas. Each row contains features for the solver's computation on each formula in the pair.

We set a threshold classification accuracy \% and report the minimum fraction of clauses
unsatisfied by the deceptively planted satisfying assignment $(u/rn)$ before our classifier can distinguish it from the original satisfiable formula with an average test accuracy \% of at least the set threshold. We start with $u = 1$ and double its value until we find a value $(U)$ that crosses the threshold classification accuracy \%. Once we find this value, we use binary search in the range $[U/2, U]$ to find the desired minimum value of $u$. For each value of $r$, $q$, and $u$, For each value of $r$, $q$, and $u$, we follow the training procedure described in \ref{expt-setup} to obtain the average test accuracy \%. We use a threshold classification accuracy \% of 70\%. This was primarily done to ensure our results are significantly better than random guessing. 

\medskip

\section{Experimental Controls for Experiment 2} \label{expt-controls}

We applied the following experimental controls:

\begin{enumerate}
    \item Fix a satisfiable formula $F$ with satisfying assignment $A$. Then for all pairs $(u, v)$ such that $u < v$, we ensure that all clauses that are not satisfied by $A$ in $F_u$ are also not satisfied in $F_v$. This is to ensure that when comparing the classification accuracy on $(F, F_u)$ and $(F, F_v)$, the only difference between the 2 pairs are the additional randomly chosen $v - u$ clauses that $A$ does not satisfy.
    \item We randomly permute the variable names and order of the clauses to ensure that the classifier cannot learn the labels or order of the clauses.
    \item When running WalkSAT on each formula in a pair, we ensure that they are initialized with the same random seed to control for differences due to randomness and starting assignment.
\end{enumerate}

\end{document}